\documentclass[conference]{IEEEtran}
\IEEEoverridecommandlockouts
\usepackage{cite}
\usepackage{amsmath,amssymb,amsfonts}
\usepackage{algorithmic}
\usepackage{graphicx}
\usepackage{textcomp}
\usepackage{xcolor}

\usepackage{comment}
\usepackage{multirow}
\usepackage{url}

\usepackage[normalem]{ulem}
\useunder{\uline}{\ul}{}

\def\BibTeX{{\rm B\kern-.05em{\sc i\kern-.025em b}\kern-.08em
    T\kern-.1667em\lower.7ex\hbox{E}\kern-.125emX}}
\begin{document}

\title{Towards Interactive Deepfake Analysis
}

\author{
    \IEEEauthorblockN{Lixiong Qin\IEEEauthorrefmark{1}\IEEEauthorrefmark{2}, Ning Jiang\IEEEauthorrefmark{1}, Yang Zhang\IEEEauthorrefmark{1}, Yuhan Qiu\IEEEauthorrefmark{1}, Dingheng Zeng\IEEEauthorrefmark{1}, Jiani Hu\IEEEauthorrefmark{2}, Weihong Deng\IEEEauthorrefmark{1}}
    \IEEEauthorblockA{\IEEEauthorrefmark{1}\textit{Mashang Consumer Finance Co., Ltd}, Chongqing, China \\
    \{lixiong.qin, ning.jiang, yang.zhang11,  yuhan.qiu, dingheng.zeng, weihong.deng\}@msxf.com}
    \IEEEauthorblockA{\IEEEauthorrefmark{2}\textit{Beijing University of Posts and Telecommunications}, Beijing, China \\
    jnhu@bupt.edu.cn}
}

\begin{comment}

\end{comment}

\maketitle

\begin{abstract}
Existing deepfake analysis methods are primarily based on discriminative models, which significantly limit their application scenarios. 
This paper aims to explore interactive deepfake analysis by performing instruction tuning on multi-modal large language models (MLLMs). 
This will face challenges such as the lack of datasets and benchmarks, and low training efficiency. To address these issues, we introduce (1) a GPT-assisted data construction process resulting in an instruction-following dataset called DFA-Instruct, (2) a benchmark named DFA-Bench, designed to comprehensively evaluate the capabilities of MLLMs in deepfake detection, deepfake classification, and artifact description, and (3) construct an interactive deepfake analysis system called DFA-GPT, as a strong baseline for the community, with the Low-Rank Adaptation (LoRA) module. The dataset and code will be made available at \url{https://github.com/lxq1000/DFA-Instruct} to facilitate further research.

\end{abstract}


\section{Introduction}

\begin{figure}[htbp]
\centering
\includegraphics[width=\columnwidth]{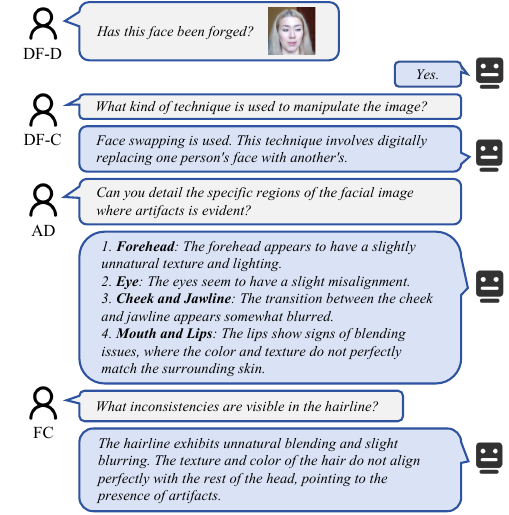}
\caption{An interactive deepfake analysis system. Questions representing the four abilities are shown separately. 
Responses are generated by our DFA-GPT.}
\label{fig:idfa}
\end{figure}

The rapid advancement of Artificial Intelligence Generated Content (AIGC) has blurred the lines between fiction and reality. 
Unauthorized deepfake images or videos can be used to manipulate public opinion, facilitate cyberbullying or extortion, and fabricate evidence. 
Deepfake analysis (DFA) is crucial for regulating and mitigating the potential negative impacts of deepfake techniques.
However, existing DFA methods \cite{FFD,MADD,SLADD,CADDM,UCF} predominantly rely on discriminative models for deepfake detection and classification, which limits their applicability.
In this paper, we aim to explore interactive deepfake analysis. 
As shown in Fig.~\ref{fig:idfa}, an interactive DFA system should first have the following fundamental capabilities: (1) deepfake detection (DF-D): determining if the input face image is forged; and (2) deepfake classification (DF-C): identifying the specific category of forgery technique used. 
Additionally, it should possess the following extended capabilities: (3) artifact description (AD): describing the artifacts present in the image that indicate forgery; and (4) free conversation (FC): answering any questions related to the forgery, including follow-up inquiries about the artifacts.
In critical domains such as social security, personal privacy protection, and judicial investigation, an interactive DFA system can alert human specialists to clues requiring further manual examination, significantly enhancing their working efficiency.

Recently, multi-modal large language models (MLLMs) \cite{GPT-4V,LLaVA,MiniGPT-4,PandaGPT,Qwen-VL,InternVL} have achieved notable success in describing and reasoning over fine-grained and complex visual cues. 
MLLMs are well-suited for sequence-to-sequence generation tasks and can thus serve as an interactive DFA system after instruction-tuning to align deepfake knowledge.
However, deploying MLLMs for interactive DFA faces 3 major challenges: (1) Lack of suitable instruction-following datasets. Existing DFA datasets \cite{data1,data2,FFpp,Celeb-DF,data3,df-40} typically only contain labels for deepfake detection and classification. Manually annotating large-scale, fine-grained 
artifact descriptions or free conversations is both labor-intensive and costly. (2) The artifact description task is currently undefined and therefore lacks evaluation metrics. (3) MLLMs usually include visual encoders and LLMs with billions of parameters, and fine-tuning the entire model can lead to prohibitively high computational costs.

To address these challenges, we first propose a GPT-assisted data construction process, capable of generating instruction-following data from real face images and videos from the internet or other available datasets. The resulting dataset, named DFA-Instruct, comprises 127.3K aligned face images and 891.6K question-answer pairs related to these images. Specifically, 127.3K question-answer pairs pertain to DF-D, 127.3K to DF-C, 127.3K to AD, and 509.7K to FC.
Based on this dataset, we introduce a new evaluation benchmark called DFA-Bench, designed to comprehensively assess the capabilities of MLLMs in deepfake analysis, which includes an evaluation of their artifact description (AD) capabilities.
Furthermore, we incorporate the Low-Rank Adaptation (LoRA) \cite{LoRA} module into MLLMs, which reduces computational costs by learning parameter residuals through low-rank matrices. 
As a result, under limited computational resources, we successfully build DFA-GPT as an interactive deepfake analysis system.
To the best of our knowledge, our study represents the first attempt for interactive deepfake analysis, exploring a new research direction for the information forensics and security community.

\section{Related Work}

In recent years, deepfake techniques have made substantial advancements. In the literature, deepfake techniques can be broadly categorized into four types: (1) face swapping (FS) \cite{SimSwap,BlendFace,UniFace,MobileSwap,e4s}, which replaces the identity of a target face with that of a source face; (2) face reenactment (FR) \cite{LIA,DaGAN,SadTalker,MCNet,HyperReenact}, which modifies source faces to imitate the actions or expressions of another face; (3) face editing (FE) \cite{e4e,StarGAN,StarGANv2,StyleCLIP}, which modifies specific facial attributes such as age, gender, hair color, etc; (4) entire face synthesis (EFS) \cite{VQGAN,StyleGAN3,SD-2.1,PixArt-α,DiT-XL/2}, which involves generating entirely new faces with GANs or diffusion models. Existing deepfake analysis methods \cite{FFD,MADD,SLADD,CADDM,UCF} typically use discriminative models to determine whether an input image is fake, but they can not provide artifact descriptions. 

Instruction-tuning is initially proposed in the field of NLP to unlock the powerful understanding and reasoning capabilities brought by pre-training to LLMs \cite{GPT-3}. 
In the past year (2023), MLLMs \cite{GPT-4V,LLaVA,MiniGPT-4,PandaGPT,Qwen-VL,InternVL} have achieved success in multi-modal content understanding and reasoning tasks. Visual instruction tuning has been introduced to MLLMs by LLaVA \cite{LLaVA}, aiming to align visual concepts with the language domain to enhance instruction-following capabilities. To efficiently deploy MLLMs for specific multi-modal content understanding and reasoning tasks, numerous parameter-efficient fine-tuning (PEFT) techniques have been developed, mainly based on P-tuning \cite{P-tuning}, adapter \cite{adapter}, and LoRA \cite{LoRA}.
In our experiments, we empirically found the effectiveness of LoRA in adapting MLLMs to interactive deepfake analysis.

\section{Dataset and benchmark}

\subsection{Data Construction Process}

\begin{figure}[htbp]
\centering
\includegraphics[width=\columnwidth]{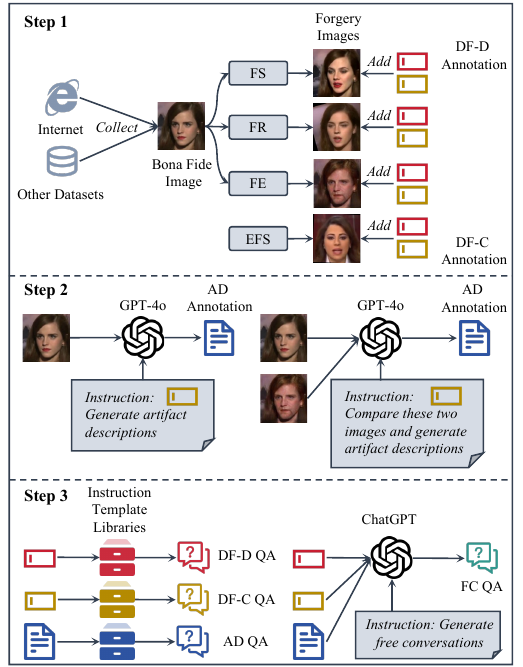}
\caption{Data construction process.}
\label{fig:datac}
\end{figure}

Our proposed data construction process for interactive deepfake analysis comprises the following three steps, as shown in Fig.~\ref{fig:datac}.

\textbf{Step 1: Acquire bona fide and corresponding forgery face images, and add annotations for DF-D and DF-C.}
Bona fide images can be sourced from the internet or existing datasets, while corresponding forgery images are generated from bona fide images using various deepfake techniques. 
Some publicly available deepfake datasets provide both bona fides and corresponding forgeries. 
However, most early datasets \cite{data1,data2,FFpp,Celeb-DF,data3} typically include only one or no more than four specific deepfake techniques, which are insufficient for training a robust interactive deepfake analysis system. 
We note that the recently released DF-40 \cite{df-40} dataset utilizes 40 distinct deepfake techniques to generate forgeries covering all four deepfake categories. 
We choose to start our data construction process with DF-40.

DF-40 uses bona fide face samples from FF++ \cite{FFpp} and Celeb-DF \cite{Celeb-DF}, while containing 40 forgery subsets, each corresponding to a specific deepfake technique. 
We first exclude subsets with too few samples. 
Among the remaining 32 subsets, only one subset belongs to the FE category, which is significantly fewer than the other categories. To address this imbalance, we replicate three face editing techniques—StarGAN \cite{StarGAN}, StarGANv2 \cite{StarGANv2}, and StyleCLIP \cite{StyleCLIP}—to generate additional forgery images.
It is noteworthy that the bona fide samples, as well as the FS and FR samples in DF-40, are stored as videos. Given the minimal changes between adjacent frames in videos, we retain frames at specific intervals. All resulting images are aligned with face alignment tools (in our implementation, we use Faceptor \cite{Faceptor}) and divided into three sets with non-overlapping identities for training, development, and testing, respectively.
Finally, we add text annotations for DF-D and DF-C to each image. Specifically, the DF-C annotation includes the name and explanation of the technique category used to generate the forgery.

\textbf{Step 2: Generate annotations for AD.}
We design two types of prompts as instructions for querying GPT-4o \cite{GPT-4o} to generate AD annotations. 
Both types of prompts include the DF-C annotations, namely the name and explanation of the deepfake technique category.
The first type of prompts requires GPT-4o to provide descriptions of the artifacts specific to certain facial regions for just a single input forgery face image, while the second type involves inputting both the forgery face image and the corresponding bona fide face image, asking GPT-4o to provide the artifacts by comparing the differences between the two images. Note that the second type of prompts is not used for image pairs from the EFS category.

\textbf{Step 3: Generate instruction-following data.}
DF-D, DF-C, and AD annotations can all be transformed into question-answer pairs just by adding instructions (questions). To enhance data diversity, all instructions are randomly sampled from instruction template libraries.
We also designed prompts to instruct ChatGPT \cite{ChatGPT} to generate conversations for the FC ability based on DF-D, DF-C, and AD annotations.

\begin{figure}[htbp]
\centering
\includegraphics[width=\columnwidth]{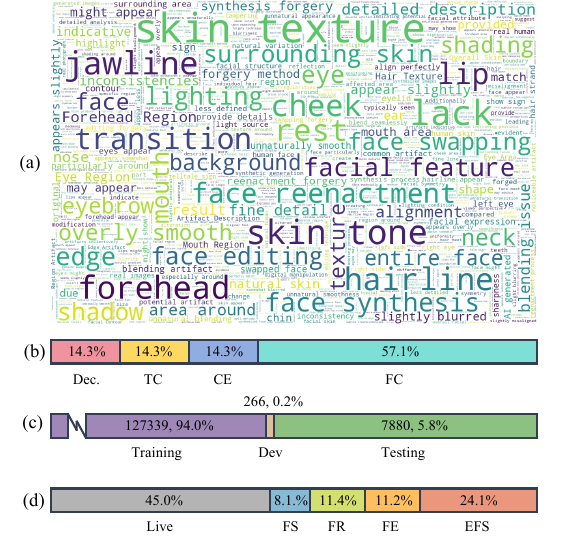}
\caption{Statistics of our proposed DFA-Instruct. }
\label{fig:data}
\end{figure}

Fig.~\ref{fig:data} provides statistics of our proposed DFA-Instruct dataset.
In Fig.~\ref{fig:data}(a), a word cloud visualizes the frequency of words within the instruction-following data, emphasizing keywords related to facial regions (e.g., skin, cheek, lip) and observation dimensions (e.g., texture, lighting, shading). These keywords indicate that artifacts have been meticulously analyzed in our proposed dataset.
Fig.~\ref{fig:data}(b) illustrates the distribution of question-answer pairs by ability; Fig.~\ref{fig:data}(c) depicts the distribution of the training, development, and testing sets; Fig.~\ref{fig:data}(d) presents the distribution of bona fide and various categories of forgery face images.

\subsection{Benchmark for Deepfake Analysis}

We develop a comprehensive metric, DFA-Bench, to evaluate the deepfake analysis capabilities of MLLMs using the test set of DFA-Instruct. 
We use ACC, ERR, and ACER to assess the DF-D ability of MLLMs, ACC to evaluate their DF-C ability, and ROUGE-L to measure their AD ability. 

To calculate these metrics, the test data need to be organized into standardized formats as follows: DF-D as assertion questions, DF-C as multiple-choice questions with five options (bona fide and four deepfake technique categories), and AD as open-ended questions. 
These formats can all be completed by adding instructions. To ensure the diversity of questions, all instructions are randomly sampled from instruction template libraries.
Notably, to ensure the validity of the evaluation, we manually review AD answers and rewrite artifact descriptions that do not match the images.

\section{Method}

\begin{figure}[htbp]
\centering
\includegraphics[width=\columnwidth]{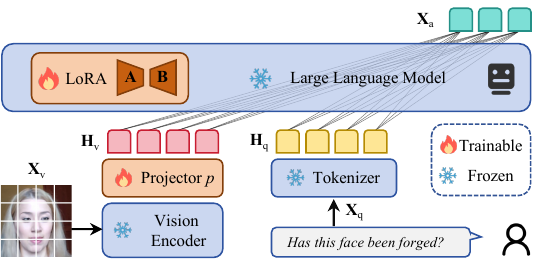}
\caption{The overall architecture of our DFA-GPT.}
\label{fig:method}
\end{figure}

The overall architecture of DFA-GPT is illustrated in Fig.~\ref{fig:method}. It consists of a vision encoder, a language tokenizer, a projector, and a large language model (LLM) with a Low-Rank Adaptation (LoRA) module.
For the input face image $\mathbf{X}_\mathrm{v}$, we use CLIP-L/14 \cite{CLIP} as the vision encoder to obtain the visual feature $\mathbf{Z}_{\mathrm{v}}=f_{\mathrm{CLIP}}(\mathbf{X}_\mathrm{v})$. 
A two-layer MLP serves as the projector, mapping $\mathbf{Z}_{\mathrm{v}}$ to the language space, resulting in $\mathbf{H}_{\mathrm{v}}=f_{p}(\mathbf{Z}_{\mathrm{v}})$. 
For the input instruction $\mathbf{X}_\mathrm{q}$, the language tokenizer converts it into language tokens $\mathbf{H}_\mathrm{q}$. 
The concatenated features of $\mathbf{H}_{\mathrm{v}}$ and $\mathbf{H}_\mathrm{q}$ are then fed into the LLM decoder. Here, we use Vicuna \cite{Vicuna} as the LLM decoder.

For efficient training, only the vision projector and the LoRA module of the LLM are learnable, while other parameters are initialized with pre-trained model weights and remain frozen. 
The LoRA module decomposes the residuals $\Delta\mathbf{W}$ of a high-dimensional parameter matrix $\mathbf{W}$ into two low-rank matrices $\mathbf{A}$ and $\mathbf{B}$, such that $\Delta\mathbf{W}=\mathbf{B}\mathbf{A}$, where the ranks of $\mathbf{A}$ and $\mathbf{B}$ are much smaller than that of $\Delta\mathbf{W}$. 
During training, only the parameters of $\mathbf{A}$ and $\mathbf{B}$ are updated. During inference, for an input $\mathbf{x}$, the output $\mathbf{h}$ is obtained by the following formula:
\begin{equation}
 \mathbf{h}=\mathbf{W}\mathbf{x} + \Delta\mathbf{W}\mathbf{x} = \mathbf{W}\mathbf{x} + \mathbf{B}\mathbf{A}\mathbf{x}.
\label{eq:LoRA}
\end{equation}

We adopt an autoregressive approach for parameter updates. 
For the input image $\mathbf{X}_\mathrm{v}$ and instruction $\mathbf{X}_\mathrm{q}$, the likelihood of the generated answer $\mathbf{X}_\mathrm{a}$ is given by:
\begin{equation}
P(\mathbf{X}_\mathrm{a} | \mathbf{X}_\mathrm{v}, \mathbf{X}_\mathrm{q}) = \prod_{i=1}^{L} p_{\theta}(x_i | \mathbf{X}_\mathrm{v}, \mathbf{X}_\mathrm{q}, \mathbf{X}_{\mathrm{a},<i})
\label{eq:likelihood}
\end{equation}
where $L$ is the length of the token sequence, $\theta$ represents the learnable parameters (including the projector parameters $p$ and the matrices $\mathbf{A}$ and $\mathbf{B}$ in the LoRA module), $x_i$ is the current token to be predicted, and $\mathbf{X}_{\mathrm{a},<i}$ denotes the previous answer tokens.

\section{Experiments}

\subsection{Implementation Details}
We initialize DFA-GPT's frozen weights with LLaVA-1.5-7B\cite{LLaVA-1.5} and only tune the projector and LoRA parameters during training. Our model is trained on DFA-Instruct for one epoch with AdamW optimizer.
The initial learning rate is set to 2e-4 and the rank of LoRA is set to 128. All experiments are conducted on 2 NVIDIA H800 GPUs.

\subsection{Comparsion With Vision-Only Models}
We compare the proposed DFA-GPT with various vision models that are well-trained. 
For a fair comparison, all vision models have their backbone parameters frozen and are tuned with only the final projection layer for one epoch respectively on the original DF-D and DF-C labels of DFA-Instruct. 
The results on DFA-Bench are shown in Table~\ref{tab:vision}.
 DFA-GPT achieves the best performance in both DF-D and DF-C. Notably, compared to the best-performing vision model, CLIP-L/14 \cite{CLIP} (which is also used as the vision encoder in LLaVA-1.5-7B \cite{LLaVA-1.5}, the initialization for DFA-GPT), DFA-GPT reduces ACER by 6.77\% in DF-D and improves accuracy by 11.23\% in DF-C. This demonstrates that introducing LLM and natural language supervision enhances the robustness of deepfake analysis systems. Additionally, DFA-GPT offers the abilities of AD and FC, which vision-only models lack.

 \begin{table}[htbp]
\caption{Comparison with vision-only models}
\begin{center}

\begin{tabular}{c|ccc|c|c}
\hline
\multirow{2}{*}{Method} & \multicolumn{3}{c|}{DF-D}                 & DF-C             & AD             \\ \cline{2-6} 
                        & ACC            & ERR           & ACER          & ACC            & ROUGE-L        \\ \hline
ResNet101\cite{ResNet}               & 74.23          & 25.77         & 25.8          & 55.84          & -              \\
DeiT-B/16\cite{DeiT}               & 75.29          & 24.71         & 23.16         & 59.86          & -              \\
DeiT-L/14\cite{DeiT}               & 78.82          & 21.18         & 20.46         & 63.54          & -              \\
CLIP-B/16\cite{CLIP}               & 81.83          & 18.17         & 17.55         & 71.12          & -              \\
CLIP-L/14\cite{CLIP}               & 88.38          & 11.62         & 11.81         & 81.51          & -              \\
\textbf{DFA-GPT (Ours)}          & \textbf{95.22} & \textbf{4.78} & \textbf{5.04} & \textbf{92.74} & \textbf{42.54} \\ \hline
\end{tabular}
\label{tab:vision}
\end{center}
\end{table}

\subsection{Ablation Study for Annotations}
As shown in Table~\ref{tab:anno}, we evaluate the impact of different type of annotations on the generalization of the deepfake analysis system.  
We note that adding DF-C annotations improves the performance of DF-D, reducing ACER by 0.87\%. Moreover, including AD annotations benefits both DF-D (ACER reduction of 0.39\%) and DF-C (ACC improvement of 0.40\%). These results indicate that more supervision signals help to build a more robust deepfake analysis system. However, incorporating free conversations can not further improve the model's performance on DF-D, DF-C, and AD. This is mainly because free conversations are derived from the previous three types of annotations and are used to enable the model to respond to free questions, without adding new information.

\begin{table}[htbp]
\caption{Ablation Study for Annotations}
\begin{center}
\begin{tabular}{c|ccc|c|c}
\hline
\multirow{2}{*}{Annotation} & \multicolumn{3}{c|}{DF-D} & DF-C    & AD      \\ \cline{2-6} 
                      & ACC       & ERR      & ACER    & ACC   & ROUGE-L \\ \hline
D                     & 94.58     & 5.42     & {\ul 5.72}    & -     & -       \\
D+C                  & 95.33     & 4.67     & {\ul 4.85}    & {\ul 92.47} & -       \\
D+C+AD               & 95.75     & 4.25     & {\ul 4.46}    & {\ul 92.87} & 42.72   \\
D+C+AD+FC            & 95.22     & 4.78     & 5.04    & 92.74 & 42.54   \\ \hline
\end{tabular}
\label{tab:anno}
\end{center}
\end{table}

\subsection{DeepFake Analysis Capability of General MLLMs}

Given that existing general-purpose MLLMs typically possess comprehensive multimodal world knowledge and excellent understanding and reasoning capabilities, we aimed to investigate whether these advanced MLLMs have sufficient deepfake analysis capabilities. We evaluate LLaVA-1.5 \cite{LLaVA-1.5}, InternVL \cite{InternVL}, and GPT-4V \cite{GPT-4V} on our DFA-Bench. The results indicate that current advanced MLLMs lack adequate understanding of face forgery. In the rapid development of AIGC, deepfake understanding should become a foundational capability for MLLMs. We hope our research will also inspire advancements in general-purpose MLLM research.

\begin{table}[htbp]
\caption{Evaluation of existing MLLMs' deepfake analysis capabilities.}
\begin{center}
\begin{tabular}{c|ccc|c|c}
\hline
\multirow{2}{*}{Method} & \multicolumn{3}{c|}{DF-D}                 & DF-C             & AD             \\ \cline{2-6} 
                        & ACC            & ERR           & ACER          & ACC            & ROUGE-L        \\ \hline
LLaVA-1.5-7B            & 54.78          & 45.22         & 43.2          & 13.95          & 13.65          \\
LLaVA-1.5-13B           & 52.16          & 47.84         & 39.92         & 11.36          & 14.00          \\
InternVL-1.2            & 53.67          & 46.33         & 41.28         & 25.57          & 9.93           \\
InternVL-1.5            & 45.48          & 54.52         & 44.92         & 16.46          & 16.40          \\
GPT-4V                  & 59.84          & 40.16         & 49.43         & 20.06          & 21.56          \\
\textbf{DFA-GPT (Ours)}          & \textbf{95.22} & \textbf{4.78} & \textbf{5.04} & \textbf{92.74} & \textbf{42.54} \\ \hline
\end{tabular}
\label{tab1}
\end{center}
\end{table}

\section{Conclusion}

This work is the first for interactive deepfake analysis. We define four key capabilities that an interactive DFA system should possess: deepfake detection (DF-D), deepfake classification (DF-C), artifact description (AD), and free conversation (FC). We aim to achieve such an interactive DFA system by instruction-tuning MLLMs. To this end, we first propose a data construction process to produce instruction-following data, resulting in DFA-Instruct. Subsequently, based on the test set of DFA-Instruct, we construct DFA-Bench, a comprehensive benchmark for evaluating MLLMs' deepfake analysis capabilities. Finally, by introducing LoRA into MLLM, we develop DFA-GPT as a strong baseline for interactive deepfake analysis. Our work provides a new research direction for the information forensics and security community.

%\section*{Acknowledgment}

%\section*{References}

\bibliographystyle{ieeetr}
\bibliography{main}

\end{document}